\documentclass[10pt,twocolumn,letterpaper]{article}

\usepackage{iccv}
\usepackage{times}
\usepackage{epsfig}
\usepackage{graphicx}
\usepackage{amsmath}
\usepackage{amssymb}
\usepackage{multirow}

\usepackage[breaklinks=true,bookmarks=false]{hyperref}

\iccvfinalcopy 

\def\iccvPaperID{****} 

\ificcvfinal\fi

\begin{document}

\title{Dual-Tuning: Joint Prototype Transfer and Structure Regularization for Compatible Feature Learning}

\author{Yan Bai$^{1,4}$, Jile Jiao$^{2}$, Shengsen Wu$^{1}$, Yihang Lou$^{1}$, Jun Liu$^{3}$,  Xuetao Feng$^{2}$, and Ling-Yu Duan$^{1,4,}$\\ 
~~~~~~~~~~$^{1}$Peking University, Beijing, China~~~~~~~~~~~$^{2}$Alibaba Group, Beijing, China \\
$^{3}$Singapore University of Technology and Design ~~$^{4}$Peng Cheng Laboratory, Shenzhen, China
\protect\\ {\tt\small \{yanbai, sswu, yihanglou, lingyu\}@pku.edu.cn,} \protect\\ {\tt\small\{jile.jjl, xuetao.fxt\}@alibaba-inc.com, jun\_liu@sutd.edu.sg}
}

\maketitle
\ificcvfinal\thispagestyle{empty}\fi

\begin{abstract}
Visual retrieval system faces frequent model update and deployment. It is a heavy workload to re-extract features of the whole database every time.
Feature compatibility enables the learned new visual features to be directly compared with the old features stored in the database. In this way, when updating the deployed model, we can bypass the inflexible and time-consuming feature re-extraction process. 
However, the old feature space that needs to be compatible is not ideal and faces the distribution discrepancy problem with the new space caused by different supervision losses. 
In this work, we propose a global optimization Dual-Tuning method to obtain feature compatibility against different networks and losses. 
A feature-level prototype loss is proposed to explicitly align two types of embedding features, by transferring global prototype information. 
Furthermore, we design a component-level mutual structural regularization to implicitly optimize the feature intrinsic structure. Experimental results on million-scale datasets demonstrate that our Dual-Tuning is able to obtain feature compatibility without sacrificing performance. (Our code will be avaliable at https://github.com/yanbai1993/Dual-Tuning)
  
\end{abstract}


\section{Introduction}
Visual matching and retrieval systems are widely utilized in many scenarios, such as image retrieval~\cite{gordo2016deep,russakovsky2015imagenet}, person re-identification (ReID)~\cite{Sun2018ECCV,Kalayeh2018CVPR,Tian2018CVPR,fu2019self}, vehicle re-identification~\cite{tang2019pamtri,lou2019veri}, and face recognition~\cite{deng2019arcface,wang2018cosface}. 
Most visual systems use deep learning models to map images into an embedding space. 
In this space, features of similar samples are close to each other, and tend to form a cluster when they share the same class. 
Commonly, the embedding features of a large-scale image collection are extracted in advance, and referred to as the gallery set or database. Then, the retrieval is accomplished by ranking all gallery features according to their distances to the input query feature. 

\begin{figure}[t]
\centering
\vspace{-10pt}
  \includegraphics[width=0.9\linewidth]{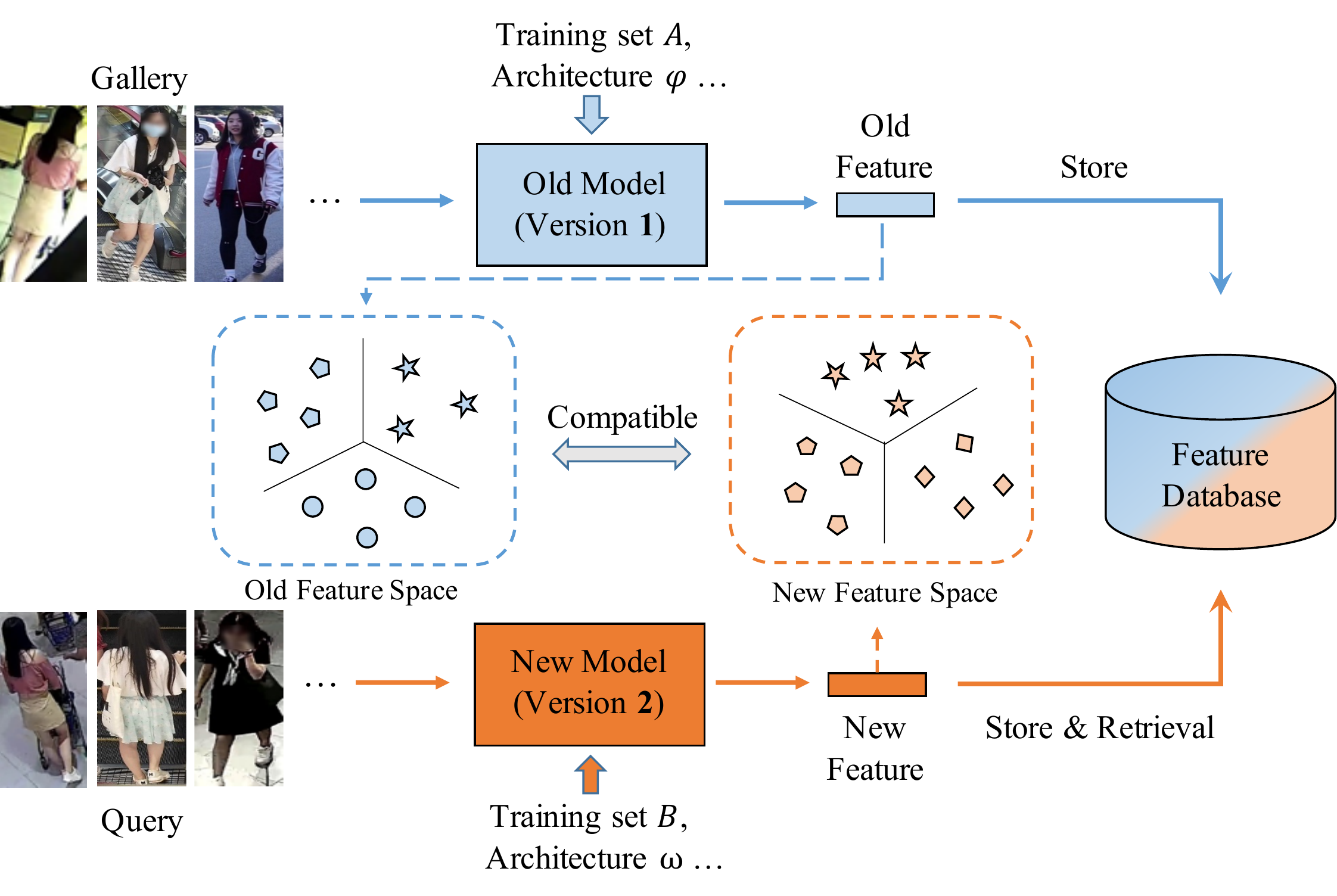}
  \caption{Illustration of feature compatible learning. }
    \label{first}
\vspace{-20pt}
\end{figure}

To achieve higher performance, the models deployed in practical industrial systems often need to be updated from time to time, due to extended training data, improved network architectures or loss functions. 
Once the model is updated, to ensure feature consistency, the entire gallery set needs to be re-extracted.
It is a time and memory consuming process because millions or even billions of gallery features have to be re-generated for a large-scale retrieval system. 
Besides, the computing power is very limited for real-time surveillance analysis. 
There is even no spare computing power for such occasional large-scale history feature re-extraction. 
Therefore, in this paper, we focus on feature compatible learning, which enables the feature extracted by the newly learned model to be directly compared to that of an old model, without sacrificing performance. 
Then the old gallery feature set can still work smoothly with the new queries, bypassing the repetitive and painful feature re-extraction process, as shown in Fig. \ref{first}. 

To achieve feature compatibility, a straightforward solution is to constrain the representations produced by the new and old models to be identical or similar. 
Budnik \textit{et.al.}~\cite{budnik2020asymmetric} introduced an asymmetric metric learning method for knowledge transfer. It projects the features of the new model into the old embedding space, and uses metric loss, such as triplet loss \cite{hadsell2006dimensionality}, to optimize the distances between new and old features. 
Besides the feature-level optimization, Shen \textit{et.al.} proposed BCT~\cite{shen2020towards}, a backward compatible training framework for feature compatible learning, which uses the classifier component of the old model to supervise the training process of the new model.

However, the compatible learning faces three challenges and the existing methods can not cope with them well. 
First, the old model's embedding space is usually not ideal and exists some sensitive limitations, such as a sparse distribution and outlier samples. 
Therefore, the behavior of imitating the disordered old representations has a potential risk of damaging the new model's ability to achieve stronger discrimination power. 
Second, when supervised by different loss functions, there exists a severe distribution discrepancy between new and old feature spaces. 
The instance-level optimization that depends on few samples is insufficient and cannot cover the whole structure of the embedding space. 
Third, for the database, the continuously added new model features should not overlap with those old model features in the embedding space. However, the existing methods ignore the optimization for the mixed dataset that have both new and old features. 
Considering the three challenges, a global optimization between the new and old embedding spaces is necessary.

Such global information can be obtained from both the embedding space and classifier's hypothesis space~\cite{zhong2020bi}. 
For dual tuning in these two spaces, 
the feature-level optimization can explicitly achieve compatibility by embedding feature alignment, and the classifier's supervision can also implicitly constrain the feature intrinsic structure to obtain compatible features. 
Therefore, it is necessary and effective to use these two types of information simultaneously, which has also been demonstrated in other fields \cite{zhong2020bi,luo2019strong}, such as the widely used triplet loss + softmax loss in ReID.

In this paper, we propose a compatible feature learning method called Dual-Tuning, including a feature-level compatible prototype loss and a component-level mutual structural regularization. 
In the embedding space, the features belonging to the same class will tend to be clustered together \cite{wen2016discriminative,luo2019strong}. 
This inspires us to represent each class as a prototype. Then the whole manifold structure can be described by the prototypes, bypassing the instance-level limitations in old embedding space.
We transfer such prototype knowledge from old embedding space to the new one. 
Thereby the new features could closely surround the old prototypes to achieve compatibility while also obtaining a more compact and discriminative space. 
To simulate the mixed gallery scenario where both new and old features exist, we further introduce a memory bank to calculate the prototypes in the new embedding space.
By utilizing both the new and old prototypes, the feature compatibility and discrimination power can be further boosted. 

Besides the feature-level optimization, a component-level mutual structural regularization is also proposed. 
Intuitively, the classifier provides the ``rules" of feature intrinsic structure.   
If features derived from two models are compatible and can mutually match, the features from one model can also satisfy the other model's rules. 
Therefore, a component interoperation scheme is designed for mutual structural regularization. 
We recombine the embedding backbone of one model and the classifier from the other model. In this way, the structural information in old classifier can be used to supervise new embedding learning. 
Simultaneously, the old embedding module can also assist in formulating the structural rules of the new classifier.

Overall, 
our contributions can be summarized as follows:
\begin{itemize}
    \vspace{-5pt}
    \item We propose a Dual-Tuning method, which exploits the global prototype in embedding space and intrinsic structure information encoded in classifier to achieve feature compatibility. 
    \vspace{-5pt}
    \item A prototype-based compatible loss is proposed, which uses prototypes to bridge and align the new and old embedding features in a global way. 
    \item A mutual structural regularization is further designed to use both the old embedding module and old classifier head for new model training. 
    
\end{itemize}
Dual-Tuning is robust against the changing factors in new model, such as network architectures, loss functions, feature dimensions, and training data. The cross-test performance between new and old features can even surpass only using the old one, without sacrificing its new feature self-test performance on million-scale datasets.

\begin{figure*}[t]
\centering
  \vspace{-5pt}
  \includegraphics[width=1\linewidth]{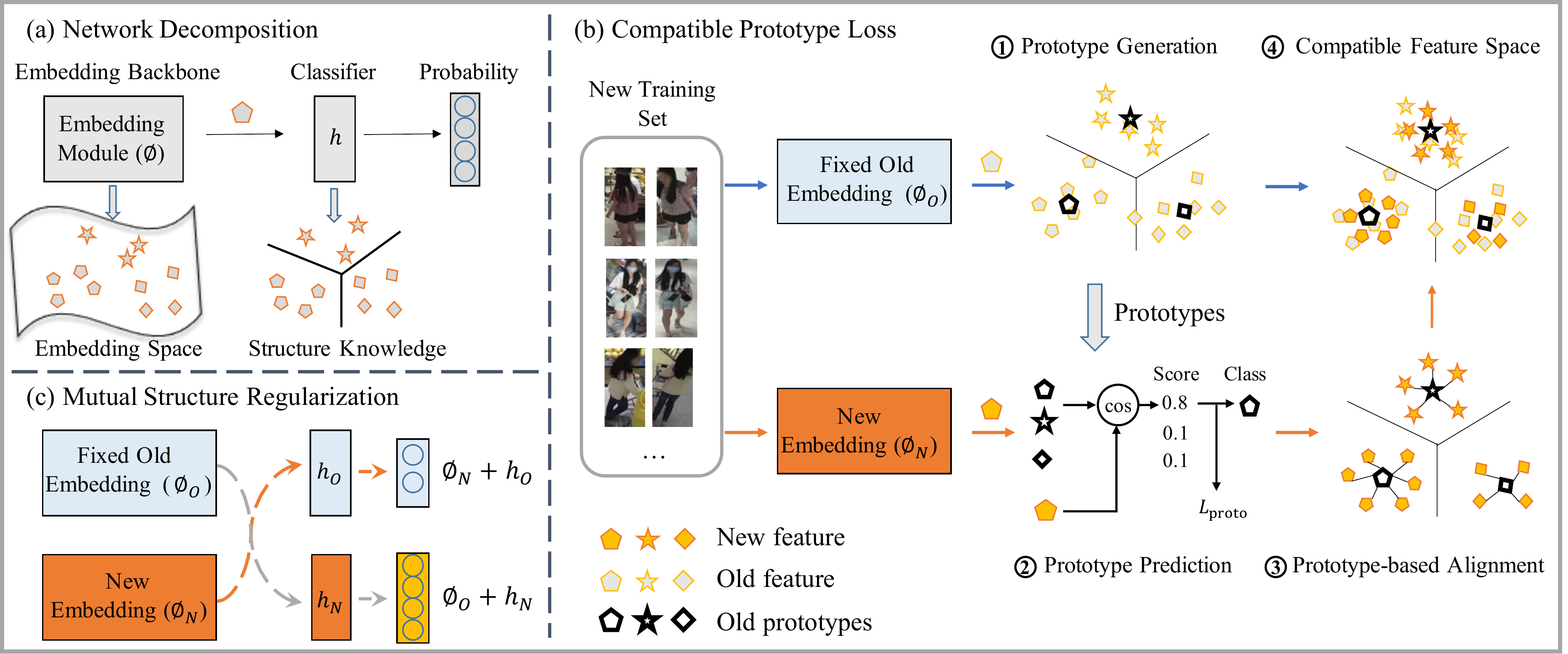}
  \vspace{-5pt}
  \caption{Illustration of the Dual-Tuning Method. (a) The network of embedding backbone + classifier head indicates the classifier will implicitly contain the intrinsic knowledge of the embedding structure. 
  (b) The compatible prototype loss uses old prototypes to bridge the old and new embedding space. By the similarity-based prototype prediction scheme, we can align the new and old features. Although these two spaces have distribution discrepancy, the new feature could closely surround old prototypes to achieve compatibility.  (c) The mutual structure regularization designs a component interoperation to constrain the new embedding backbone and classifier simultaneously.
  } 
  \vspace{-10pt}
    \label{framework}
\end{figure*}

\section{Related Work}
\subsection{Feature Compatible Learning}

Due to the increasing size of gallery sets and the heavy workload of re-extracting gallery features, feature compatible learning was proposed recently.
Previous works \cite{wang2020unified,hu2019towards} aimed to learn a separate mapping model to translate features from a source model to a target model. However, an additional feature re-representation process is needed before the feature can be compared with the stored gallery features. 
Recently, Shen \textit{et al.} \cite{shen2020towards} proposed BCT, a backward-compatible training framework with a regularization function by using the old classifier in training the new embedding model, without additional feature re-generation process.
However, when the classifiers/loss functions of new and old models are different, 
BCT faces severe performance drops.
Differently, our Dual-Tuning can explicitly align embedding features by transferring the prototype information, and implicitly constrain the feature intrinsic structure in a bi-directional manner.
In this way, we can better achieve feature compatibility even the old embedding space is not ideal and has distribution discrepancy with the new space. 

\subsection{Domain Adaptation}
Domain adaptation~\cite{xu2020cross,wang2020prototype} aims to learn a model for the unlabeled target domain by leveraging knowledge from a labeled source domain. A group of methods uses various metrics to mitigate the distribution difference caused by domain gap, such as maximum mean discrepancy (MMD) \cite{fukumizu2009kernel} and its variants \cite{yan2017mind, shen2018wasserstein}. 
Another typical line aims to design a feature space such that confusion between the source and target distributions in that space is maximal \cite{ganin2015unsupervised,pei2018multi,zou2018unsupervised}.
These methods can be used to align the (marginal) distribution of the new and old classes.  The problem we focus on differs in that we aim to achieve feature-level compatibility between any feature pair of new and old models.

\subsection{Knowledge Distillation}
Knowledge distillation aims to guide the learning of a small network using the knowledge from a large network. 
Hinton \textit{et al.} \cite{hinton2015distilling} proposed using the probabilistic outputs of a larger network as soft targets to supervise the smaller network, and KL Divergence loss is adopted. 
Besides, some variants were proposed to mimic the intermediate feature maps~\cite{yao2020knowledge}, attention maps~\cite{zagoruyko2016paying}, or second-order statistics (Gram matrix)~\cite{yim2017gift} from the teacher network. 
However, knowledge distillation is not applicable in compatible learning. Since the existing teacher model (old model) is unsatisfactory, mimicking a ``weak'' teacher will affect new models to get better discrimination capability. 

\section{Proposed Method}


\subsection{Problem Formulation}
In this work, we focus on feature compatible learning, which requires the learned features of a new (updated) model to be directly compared with features of its old version. 
Typically, a network model can be split into two components, an embedding backbone $\phi(\cdot ;\omega_\phi)$ and a task head $h(\cdot ;\omega_h)$, where $\{\omega_\phi, \omega_h\}$ are the network parameters. 
For simplicity, those two components are also denoted as $\phi(\cdot)$ and $h(\cdot)$.  
In visual retrieval, the classification network is widely used to obtain the embedding feature \cite{luo2019strong}.
The embedding module $\phi: \mathcal{X}\rightarrow \mathcal{F}$ maps an input $x \in \mathcal{X}$ to an embedding space $\mathcal{F}$, 
and the head module $h: \mathcal{F}\rightarrow \mathcal{P}$ further projects the features to the classifier's hypothesis space $\mathcal{P}$. 
Then the category probabilities can be obtained by $p = h(\phi(x; \omega_{\phi});\omega_{h})$, as shown in Fig. \ref{framework}(a). 
In feature compatible learning, the old model can be represented as $\phi_O$ and $h_O$ trained on the old training set $\tau_{O}$. 
As model structure evolves or data increases, a new model with $\phi_N$ and $h_N$ will be generated based on the new training set $\tau_{N}$. In $\tau_{N}$, the overlapping classes with $\tau_{O}$ is denoted as $\tau_{N'}$. 

After model updating, some of the gallery features stored in the database are still derived from the old model, while the queries are extracted from the new model. 
The compatible feature learning makes it possible to conduct retrieval between the new query features and old gallery features. 
The expected compatible features derived from the new model should satisfy two criteria. First, it should be compatible with the old features. 
Second, compatible learning should not affect the new model optimization to get better performance from updated data or network designs.


The distance relationship in the compatible embedding space can be described as, 
\begin{equation}
\begin{aligned}
 d(&\phi_{N}(a), \phi_{*}(p)) < d(\phi_{N}(a), \phi_{*}(n)),  \phi_* \in \{\phi_N, \phi_O\},\\
&  \forall(a,p,n)\in \{(a,p,n)|y_a = y_p \  \text{and} \  y_a \neq y_n\}, 
\label{eq:new}
\end{aligned}
\end{equation}
where $a$ denotes the query sample, and $p$, $n$ denote positive and negative galleries. 
$y$ is the class label, and $d(.)$ is the Euclidean distance. This formulation can be described as, no matter if the gallery features are from the old or new embedding space, the distance between the positive pair $(a, p)$ should be consistently smaller than the negative pair $(a, n)$.


\subsection{Compatible Prototype Loss}

\subsubsection{Center-based Prototype Loss}
To enable feature compatibility, we design a center-based prototype loss to achieve a global optimization between new and old embedding spaces, as shown in Fig. \ref{framework} (b).
Considering the old model features are fixed and will not change during the new model's training, we use prototypes \cite{snell2017prototypical, wen2016discriminative} to represent the manifold of the old embedding space. 
By transferring such manifold knowledge, 
the new embedding space can be embedded into the old one to achieve feature alignment explicitly. 
Note that, for each sample $x$ in the \textbf{new} training set $\tau_{N}$, we first calculate its fixed old embedding feature $\phi_{O}(x)$. 
Then the prototype for each class can be represented by old feature centers as,
\begin{equation}
m_o^c = \frac{1}{|\mathcal P(c)|}\sum_{\substack{x \in \mathcal P(c)}} \phi_{O}(x),
\end{equation}
where $\mathcal P(c)$ is the sample set of class $c$, and $|\mathcal P(c)|$ is the sample number of $\mathcal P(c)$. 
In this way, given a newly added class not contained in the old training set, we can also calculate its old feature center by $\phi_{O}$. 
For $C$ classes in new training set $\tau_{N}$, we can obtain $C$ centers $\mathcal M_O = \{m_o^c\}_{c=1}^{C}$. 
These centers are robust for describing the whole embedding space. We did a toy experiment to validate it. A ReID model on Market1501 has only 61.49\% mAP, yet replacing query as its class center can get 84.52 mAP. 


To use all of the old prototypes for global supervision, we design a prototype-based prediction strategy. 
Instead of incorporating an extra classifier for feature classification, our prediction is built upon the similarities to achieve an explicit distance optimization.  
Concretely, we select a query $\phi_{N}(x)$ with new feature and all prototypes $\mathcal M_O$ from old features, then compute the cosine similarities between them.
Based on the similarity scores, we aim to predict which prototype/class the query belongs to. Thus, the constraint of prototype loss $\mathcal {L}_\text{proto}$ can be formulated as, 
\begin{equation}
 \mathcal L_\text{proto}(\phi_{N};\mathcal M_O, \tau_{N})  = - 
 \mathrm{log} \frac{\mathrm{exp}(\langle \phi_{N}(x^c), m_o^c\rangle)}{\sum_{c'=1}^C \mathrm{exp}( \langle \phi_{N}(x^c), m_o^{c'}\rangle)}, 
\label{eq:proto}
\end{equation}
where $\langle \cdot, \cdot \rangle$ denotes the cosine similarity. $x_c \in \tau_{N}$ belongs to the same class $c$ as $m_o^c$. 
It is worth mentioning that the proposed $\mathcal L_\text{proto}$ shows the following major advantages. Prototype loss is able to maximize the similarity between $\phi_{N}(x_c)$ and $m_o^c$, meanwhile minimize the similarities to \textit{\textbf{all}} other old prototypes (\textit{i.e.}, towards 0). 

\begin{figure}[t]
\centering
  \vspace{-5pt}
  \includegraphics[width=1\linewidth]{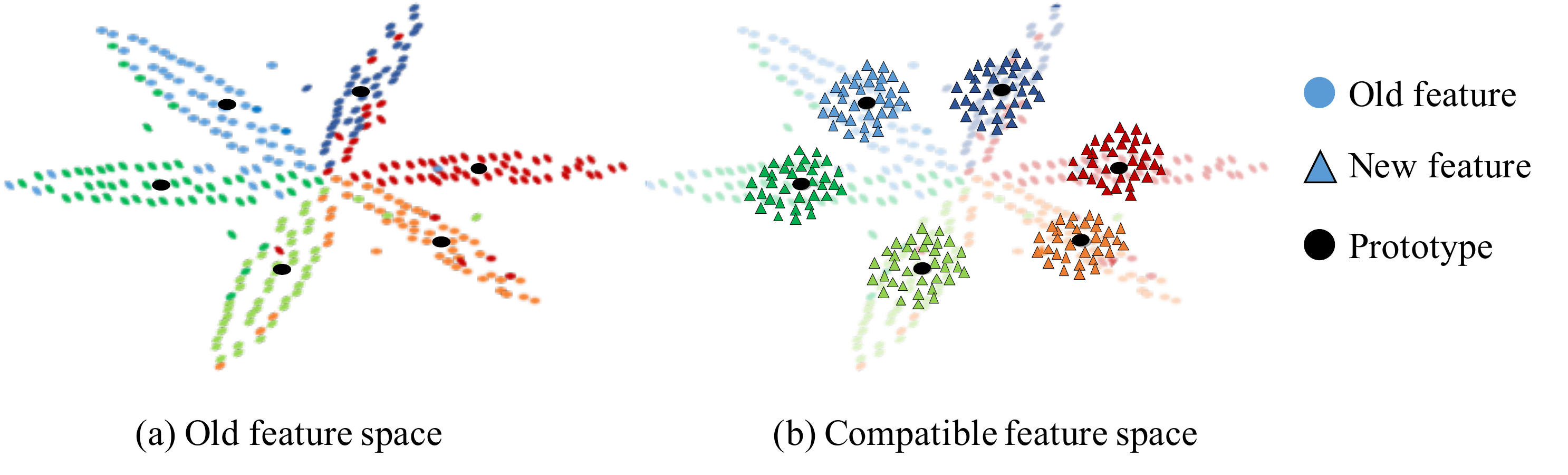}
  \vspace{-5pt}
  \caption{Illustration of the compatible feature space.
  } 
  \vspace{-10pt}
    \label{feature_space}
\end{figure}

1) Prototype loss can be treated as a similarity-based classification. Different from other classification loss, it can explicitly optimize the similarity in embedding space. 
Besides, all old prototypes involve the computation, which can bridge the new and old embedding space, and well support the global embedding optimization. 

2) The old embedding space may have outlier, and face distribution discrepancy with new space caused by different supervision losses. 
Using the prototype rather than instance feature or classifier can alleviate these limitations. As shown in Fig. \ref{feature_space}, the new features will closely surround the old prototypes to achieve compatibility while also pursuing a better representation.  

3) The compatible prototype loss is not limited to the assumption in previous work \cite{shen2020towards} that requires the old and new training data to have overlap.
It can work well with only the new dataset $\tau_{N}$ and old embedding backbone $\phi_{O}$. 


\subsubsection{Memory-based Prototype Loss}
Center-based prototype loss simulates the scene where the new query matches the old gallery, \textit{i.e.}, $d(\phi_{N}(a), \phi_{O}(p)) < d(\phi_{N}(a), \phi_{O}(n))$ in Eq. \ref{eq:new}.  However, besides the existing old features, the new features will also be continuously added into the datasbase, after obtaining the new model. 
Therefore, $d(\phi_{N}(a), \phi_{N}(p)) < d(\phi_{N}(a), \phi_{O}(n))$ and $d(\phi_{N}(a), \phi_{O}(p)) < d(\phi_{N}(a), \phi_{N}(n))$ in Eq. \ref{eq:new} should also be satisfied. 

To achieve this, we further design a memory-based prototype loss in the new embedding space. 
Since the new model is continuously updated during training, we cannot compute the whole dataset features to obtain the centers in each iteration. 
Therefore, we use a dynamic memory bank \cite{wu2018unsupervised, he2020momentum} to store the embedding features.
The memory bank is implemented by a queue, in which the current mini-batch enqueued and the oldest mini-batch dequeued. The queue can decouple the calculated embedding number from the mini-batch size. 
Then we design a discrete dictionary to store the class ID and corresponding prototypes. For the $C$ classes in the current queue, the memory $\mathcal M_N$ has $C$ key-value pairs to store the class ID $c$ and corresponding prototypes $m_n^c$.  
At each iteration, we calculate the prototypes with the features in the current queue as,
\begin{equation}
m_n^c = \frac{1}{|\mathcal Q(c)|}\sum_{\substack{x \in \mathcal Q(c)}} \phi_{N}(x),
\end{equation}
where $\mathcal{Q}_c$ denotes the samples from class $c$ in current queue. 

By considering both the new and old prototypes, we can simulate the mixed gallery situations. For each class, we randomly select its new and old prototypes, \textit{i.e.}, $m^c \in \{m_o^c, m_n^c\}$ and $\mathcal M = \{\mathcal M_O, \mathcal M_N\}$.
Then, updating Eq.~\ref{eq:proto}, the final compatible prototype loss can be represented as, 
\begin{equation}
 \mathcal L_\text{proto}(\phi_{N};\mathcal M, \tau_{N})  = - \mathrm{log} \frac{\mathrm{exp}(\langle \phi_{N}(x^c), m^c\rangle)}{\sum_{c'=1}^C \mathrm{exp}( \langle \phi_{N}(x^c), m^{c'}\rangle)}. 
\label{eq:proto2}
\end{equation}

\subsection{Mutual Structural Regularization}
Besides the feature-level optimization, we also design a component-level mutual structural regularization to further improve the compatibility, as shown in Fig. \ref{framework}(c).
Works\cite{zhong2020bi,luo2019strong} in other fields demonstrate it is effective to use embedding and classification spaces simultaneously(e.g., triplet+softmax). Therefore, if the head is available, our mutual structural regularization can further provide knowledge in classifier to get more enhanced results.

Given a network, it can be split into two components, \textit{i.e.}, an embedding module for feature extraction and a head for target task. 
Such a head can be treated as the ``rules", which contains the intrinsic structure of embedding space. 
Take a classifier head as an example, classifier head formulates classification rule, and according to this rule, the embedding in feature space $\mathcal F$ can be mapped to category probabilities in classifier's hypothesis space $\mathcal P$, $h_{\omega_{h}}: \mathcal F \rightarrow \mathcal P$. 

Therefore, if features from two models can conduct mutual matching, the features derived from one model can also produce good predictions when passing through the other model's classifier. 
This inspires us to design a component interoperation for mutual structural regularization. 
Specifically, the distribution structure of the new features should satisfy the ``rules" of the old classifier, and correspondingly, the new classifier should also predict correctly based on the old features. 

For a classification task, the widely used cross-entropy loss can be expressed as,
\begin{equation}
    \mathcal{L}_{\text{CE}}({\phi}, {h}; \tau) = \sum_{(x_i, y_i) \in \tau} - y_i log \ h(\phi(x_i; \omega_{\phi});\omega_{h} ),
\end{equation}
where $y_i$ is the one hot vector label of sample $x_i$. For training a single new or old model, the optimization targets are $\mathcal{L}_{\text{CE}}({\phi_N}, {h_N}; \tau_N)$ and $\mathcal{L}_{\text{CE}}({\phi_O}, {h_O}; \tau_O)$, correspondingly. 

The component interoperation means that, for the new and old models, we can recombine the embedding module of one model and classifier head of the other model while still maintaining good prediction performance. 
Thus, the optimization targets of mutual structural regularization $\mathcal{L}_{\text{stru}}({\phi_N}, {h_N}; \tau_N, \tau_{N'})$ can be expressed as,
\begin{equation}
    arg \min_{\omega_{\phi_N}, \omega_{h_N}}(\mathcal{L}_{\text{CE}}({\phi_N}, {h_O}; \tau_{N'}) + \mathcal{L}_{\text{CE}}({\phi_O}, {h_N}; \tau_N)).
  \label{eq:stru}
\end{equation}
Different from BCT \cite{shen2020towards} which only uses the old classifier head to supervise the new embedding module, our component interoperation can provide dual supervision. The old ``rules" will guide the training of the new embedding module. At the same time, the old features will also assist in formulating the embedding rules of the new classifier. 

\subsection{Overall Objective}
Finally, to simultaneously exploit the prototype knowledge in the embedding space and conduct mutual structural regularization by old components, the final optimization objective for Dual-Tuning is,
\begin{equation}
\begin{aligned}
    \mathcal L_{\text{all}} =  \mathcal L_{\text{proto}}(\phi_{N};\mathcal M, \tau_{N}) +  \mathcal{L}_{\text{stru}}({\phi_N}, {h_N}; \tau_N, \tau_{N'}) \\
    + \mathcal L_{\text{CE}}({\phi_N}, {h_N}; \tau_N),
\end{aligned}
\label{eq:all}
\end{equation}
where $\mathcal L_{\text{CE}}$ is the supervision loss for the target task, which is not limited to cross-entropy loss and does not need to be the same form as the old model. 
$\mathcal L_{\text{proto}}$ and $\mathcal{L}_{\text{stru}}$ are illustrated in Eq.~\ref{eq:proto2} and Eq.~\ref{eq:stru}, respectively.

\section{Experiments}
\subsection{Experiment Setting}

\textbf{Dataset.}
To comprehensively evaluate the compatible learning approaches, besides two widely used large scale person ReID datasets Market1501~\cite{zheng2015scalable} and MSMT17~\cite{Wei2017Person}, we also conduct experiments on two million-scale classification datasets ImageNet~\cite{deng2009ImageNet} and Place365~\cite{zhou2017places} used in \cite{shen2020towards}. 
We evaluate the compatible retrieval performance on the validation or testing sets of these four datasets. For Market1501 and MSMT17, the standard person ReID testing process is performed~\cite{Wei2017Person}. For ImageNet and Place365, each image is considered as a query image, and all other images are considered as gallery images, the same setting as in \cite{shen2020towards}. 

\begin{table}[!h] 
\vspace{-8pt}
\caption{Datasets information. }
\vspace{-8pt}
\label{tab:dataset}
\centering \footnotesize
\renewcommand\arraystretch{1}
\resizebox{0.9\columnwidth}{!}{
\begin{tabular}{c c c c  c c } \\
\hline
 Datasets & ImageNet & Place365 & Maket1501 & MSMT17 \\ \hline
 Classes & 1,000 & 365 & 1,501 & 4,101 \\
 Images & 1.2 million & 1.8 million & 32,886 & 126,411 \\ \hline
 \end{tabular}}
 \vspace{-8pt}
\end{table}

\newcommand{\tabincell}[2]{\begin{tabular}{@{}#1@{}}#2\end{tabular}}
\begin{table*}[t] 
\vspace{-5pt}
\caption{Performance comparison on the MSMT17 and Market1501 datasets. For a good compatible model, its cross-test performance should be comparable or even better to the old model, and self-test performance should be as good as the independently trained new model.  }
\vspace{-5pt}
\label{tab:market-result}
\centering \footnotesize
\renewcommand\arraystretch{1}
\resizebox{2.1\columnwidth}{!}{
\begin{tabular}{c c l c c c l c cc l c c c l c cc } \\
\hline
  & && \multicolumn{7}{c}{MSMT17} &&  \multicolumn{7}{c}{Market1501} \\  \cline{4-10} \cline{12-18} 
 Settings  & && \multicolumn{3}{c}{Cross-test} &&  \multicolumn{3}{c}{Self-test} && \multicolumn{3}{c}{Cross-test} &&  \multicolumn{3}{c}{Self-test}  \\ 
(new v.s. old) & Method && mAP& Top-1  & Top-5 && mAP& Top-1   & Top-5 && mAP& Top-1   & Top-5 &&mAP& Top-1   & Top-5 \\ \cline{1-2} \cline{4-10} \cline{12-18} 
\multicolumn{18}{c}{(a) ResNet18 for new model and ResNet18 for old model} \\ \hline
 Res18-50\% &Independent (old) &&  -&- & - && 28.32 & 55.97 & 70.78 &&  -&- & - && 61.49 & 82.36 & 92.52  \\
Res18-100\% & Independent (new) && 4.15 & 10.38 & 25.64 && 42.01 & 68.91 & 81.89 && 13.55	&  21.35 & 42.19 && 80.43 & 92.31 & 97.54 \\ \hline 
\multirow{5}{*}{\begin{tabular}[c]{@{}c@{}} \\ Res18-100\% \\ v.s. \\ Res18-50\%  \end{tabular} }  & $L2$ Loss~\cite{yu2019learning} && 9.87 & 20.32 & 39.73 && 36.61 & 63.28 & 79.43 && 13.35 & 24.6 & 43.12  && 73.24 & 88.97 & 94.21  \\
& KL Divergence~\cite{hinton2015distilling}  && 29.34 & 56.31 & 72.47  && 37.17 & 64.62 & 78.87 && 64.56 & 83.15 & 93.18 && 73.90 & 89.20 & 95.64   \\
& BCT~\cite{shen2020towards} && 30.12 & 58.83 & 74.93 && 39.95 & 65.57 & 80.54   && 66.28 & 85.45 & 94.60  && 77.59 & 91.27 & 97.18  \\ 
& Asymmetric Triplet~\cite{budnik2020asymmetric}  && 26.60 & 54.29 & 69.92 && 40.74 & 68.32 & 81.04 && 65.86 & 83.85 & 94.12 && 79.67 & 91.28 & 96.97\\ 
& Compatible Prototype (Ours) &&  31.11 & 58.85 & 75.99 && \textbf{42.58} & 69.59 & 82.76 && 68.12 & 86.10 & 94.85 && 80.72 & \textbf{92.30} & 97.21  \\
& Dual-Tuning (Ours) && \textbf{32.08} & \textbf{59.70} & \textbf{76.09} && 42.17 & \textbf{70.06} & \textbf{82.83 } &&  \textbf{69.53} & \textbf{86.67} & \textbf{95.04}  && \textbf{80.75} & 92.04 & \textbf{97.33} \\ \hline
\multicolumn{18}{c}{(b) ResNet50 for new model (2048-dim feature) and ResNet18 for old model (512-dim feature)} \\ \hline
Res18-50\% & Independent (old)  &&  - & -  & -  && 28.32 & 55.97 & 70.78 & &  -&- & - && 61.49 & 82.36 & 92.52   \\
Res50-100\% & Independent (new) && 0.18 & 0.04 & 0.31 && 48.23 & 73.07 & 84.57 && 0.24 & 0.06 & 0.71 && 85.14 & 94.09 & 98.13   \\ \hline
\multirow{5}{*}{\begin{tabular}[c]{@{}c@{}} \\ Res50-100\% \\ v.s. \\ Res18-50\%  \end{tabular} } 
& $L2$ Loss~\cite{yu2019learning} && 6.91 & 18.37 & 37.21 && 42.84 & 69.38 & 81.47 && 9.54 & 20.42 & 31.72 && 79.76 & 92.39 & 97.19 \\
& KL Divergence~\cite{hinton2015distilling} && 29.73 & 59.07 & 74.43 && 42.28 & 69.04 & 81.23 && 65.69 & 85.21 & 94.51  && 79.34 & 92.46 & 97.74  \\ 
& BCT~\cite{shen2020towards} && 30.83 & 60.24 & 77.01 && 45.80 & 71.76 & 83.77 && 67.45 & 85.93 & 95.07  && 82.68 & 93.14 & 97.77  \\ 
& Asymmetric Triplet~\cite{budnik2020asymmetric}  && 26.77 &  55.32 & 71.63 && 46.59 & 72.39 & 84.48 && 66.92 & 83.94 & 94.18  && 84.07 & 93.53 & 98.13 \\ 
& Compatible Prototype (Ours) && 32.18 & 60.06 & 76.68 && 47.71& 73.33 & 84.96 && 68.47 & 86.52 & 95.31 && 85.22 & 94.24 & 98.04  \\
& Dual-Tuning (Ours) && \textbf{33.85} & \textbf{61.91} & \textbf{78.43} &&  \textbf{48.77} & \textbf{74.13} & \textbf{85.32} &&  \textbf{69.98} & \textbf{86.74} & \textbf{95.40}  && \textbf{86.41} & \textbf{94.42} & \textbf{98.15} \\ \hline
 \end{tabular}
 }
\vspace{-15pt}
\end{table*}

\textbf{Implementation details.}
For compatible model learning, we adopt two widely used architectures ResNet18 and ResNet50~\cite{he2016deep} as our backbones, whose feature dimensions are 512 and 2048 respectively. For Market1501 and MSMT17, we employ the bag-of-tricks scheme \cite{luo2019strong} for ReID model training, and the input size is 256$\times$128. For ImageNet and Place365, we adopt the training scheme in \cite{shen2020towards}, and the image input size is 224$\times$224. 
For the memory bank, the queue size is set to 4096. 
The supervision loss ($\mathcal L_{\text{CE}}$ in Eq. \ref{eq:all}) for target task is cross-entropy loss. 
We optimize the model with Adam optimizer and use 1 GPU for network training. The model is trained for 120 epochs with a start learning rate of $3.5\times10^{-4}$ and performs learning rate decay $1/10$ in the $40{th}$ and $70{th}$ epochs.

\textbf{Experiment setting.}
We conduct ``\textbf{Cross-test}" and ``\textbf{Self-test}" experiments for compatible model evaluation: 
\begin{itemize}
    \vspace{-5pt}
    \item Cross-test: In retrieval, the query and gallery features are extracted from the new and old model respectively, and the Euclidean distances are directly computed between the new and old features. If the dimensions of these two types features are different, zero padding will be used for dimension alignment \cite{shen2020towards}. 
    \vspace{-5pt}
    \item Self-test: In retrieval, the query and gallery features are both extracted from the new (learned) model. 
    \vspace{-5pt}
\end{itemize}
The mean Average Precision (mAP) and Top-K accuracy are used as evaluation metrics. 

To emulate the practical case that a new model is generated when a new larger training set is available,
half (50\%) of the classes are used for training an old model, and all (100\%) classes are used for training a new model. 
In our table presentation, the top part shows the performances of the independent models. 
The bottom part provides the performances of feature compatibility learning. 



\subsection{Compatibility Analysis}
The results of Market1501, MSMT17, Place365, and ImageNet are shown in Table \ref{tab:market-result}, \ref{tab:Place365-result}, and  \ref{tab:ImageNet-result}.  
The ``Res18 v.s. Res18" or ``Res50 v.s. Res18" in setting (new v.s. old) column means the new model is with ResNet18 / ResNet50 backbone, and the old model is ResNet18. 
Several comparison methods are adopted, including a naive $\mathcal L2$ loss~\cite{yu2019learning}, knowledge distillation method KL Divergence~\cite{hinton2015distilling}, metric-based Asymmetric triplet loss~\cite{budnik2020asymmetric}, and compatible learning method BCT~\cite{shen2020towards}. 

\textbf{Independently trained new and old model. } We first conduct a simple experiment to test the cross-test performance between two independently trained models, \textit{i.e.}, Independent (new) and Independent (old) in Table \ref{tab:market-result}. 
The cross-test performance between these two independent models is almost 0\% accuracy in most experiments, demonstrating its bad compatibility. In addition, the cross-test between Res18-50\% and Res18-100\% models on Market1501 is 13.55\% mAP, since they use the same initialization parameters, which is helpful for feature compatibility. 

\textbf{A naive $\mathcal L2$ loss. } 
Given a sample, the $\mathcal L2$ loss aims to minimize the Euclidean distance between its new and old features \cite{yu2019learning}. 
However, such a straightforward approach is too local and strict for new feature learning.  It not only fails to achieve feature compatibility, but also interferes with the training of new models, as shown in Table \ref{tab:market-result} and \ref{tab:ImageNet-result}. 

\textbf{Knowledge distillation - KL Divergence. } 
We also compare with the classic KL Divergence \cite{hinton2015distilling} for knowledge distillation, which constrains the probabilistic outputs of the new model by those of the old model. 
However, the new model (student) should achieve better performance than old model (teacher). Therefore, using  the ``bad" teacher to guide the learning of a new model will affect obtaining a better embedding space. As shown in Table \ref{tab:market-result}(a), the independently trained ResNet18 model obtains 80.43\% mAP on Market1501, while supervised by KL Divergence, it only obtains 73.90\% mAP on self-test.

\textbf{Metric-based asymmetric triplet loss. } 
For the asymmetric triplet, we adopt the triplet loss to optimize the distances between new and old features, and the triplet unit is $(\phi_N(a), \phi_O(p), \phi_O(n))$. Such asymmetric triplet is first used for knowledge transfer learning in \cite{budnik2020asymmetric}.
On the MSMT17 dataset (Table \ref{tab:market-result}(a)), the asymmetric triplet loss can only achieve 26.60\% cross-test mAP. Such performance illustrates that when the scale of the dataset becomes larger, our prototype-based global optimization can achieve significant advantages over pair-based local optimization. 

\textbf{Backward-compatible learning - BCT. }
BCT~\cite{shen2020towards} uses the old classifier head to supervise the new embedding module to achieve feature compatibility. 
BCT and Asymmetric triplet loss achieve comparable performances. For example, on the Market1501 datasets (Table \ref{tab:market-result}(a)), the asymmetric triplet loss achieves better performance on self-test (79.67\% v.s. 77.59\% mAP), and slightly worse performance on cross-test, while on the Place365 and ImageNet datasets (Table  \ref{tab:Place365-result}, \ref{tab:ImageNet-result}), opposite results are obtained. 

\vspace{5pt}
By comparison, our Dual-Tuning achieves the best performances on all datasets.
Impressively, the cross-test performance of our Dual-Tuning beats the old model by a remarkable margin, \textit{e.g.}, on Market1501 Res18 v.s. Res18 setting, 8.04\% mAP improvement can be obtained (69.53\% v.s. 61.49\%). Such a performance gain indicates that the prototype loss is able to push the new query features close to the old feature centers of the same ID. 
Besides, on the Market1501, MSMT17, and Place365 datasets, the self-test performances even surpass the performance of independently trained new model, demonstrating that the old model information is also useful for new model learning. 

\begin{table}[t] 
\vspace{-5pt}
\caption{Performance comparison on the Place365 dataset. }
\vspace{-5pt}
\label{tab:Place365-result}
\centering \footnotesize
\renewcommand\arraystretch{1}
\resizebox{1\columnwidth}{!}{
\begin{tabular}{c c c c  c c } \\
\hline
 Settings   && \multicolumn{2}{c}{Cross-test} & \multicolumn{2}{c}{Self-test}  \\ 
 (new v.s. old) & Method  & Top-1   & Top-5 & Top-1   & Top-5 \\ \hline 
Res18-50\% & Independent (old) &  - & - &27.0 & 55.9 \\
Res18-100\%  & Independent (new) & 0.0 & 0.2 & 33.0 & 62.0  \\ \hline
\multirow{3}{*}{\begin{tabular}[c]{@{}c@{}} Res18-100\% \\ v.s. \\ Res18-50\% \end{tabular}} &Asymmetric Triplet~\cite{budnik2020asymmetric} &  29.4 & 58.2 & 29.1 & 58.1 \\
&BCT~\cite{shen2020towards}  & 27.3 & 57.8 & 31.9 & 62.2 \\ 
& Dual-Tuning (Ours) &   \textbf{31.2} &  \textbf{60.8}  &  \textbf{33.7} &  \textbf{62.6} \\ \hline \hline
Res18-50\%  &Independent (old)  &  - & - & 27.0 & 55.9 \\
Res50-100\%  &Independent (new) &  0.0 & 0.1 & 35.1 & 64.0   \\ \hline
\multirow{2}{*}{\begin{tabular}[c]{@{}c@{}} Res50-100\% \\ v.s. \\ Res18-50\% \end{tabular}} 
&Asymmetric Triplet~\cite{budnik2020asymmetric} & 29.5 & 58.7 & 31.0 & 60.8\\
& BCT~\cite{shen2020towards} & 27.5 & 57.8 & 32.9 & 62.2 \\ 
& Dual-Tuning (Ours) &  \textbf{28.5} & \textbf{58.0}  & \textbf{35.2} & \textbf{64.0} \\ \hline
\end{tabular}
 }
 \vspace{-5pt}
\end{table}

\begin{table}[t] 
\caption{Performance comparison on the ImageNet dataset. }
\vspace{-5pt}
\label{tab:ImageNet-result}
\centering \footnotesize
\renewcommand\arraystretch{1}
\resizebox{1\columnwidth}{!}{
\begin{tabular}{c c  c c c c } \\
\hline
 Settings  && \multicolumn{2}{c}{Cross-test} &  \multicolumn{2}{c}{Self-test} \\ 
(new v.s. old) & Method & Top-1   & Top-5 & Top-1   & Top-5 \\ \hline 
 Res18-50\% & Independent (old) & - & - & 39.5 & 60.0  \\
Res50-100\%  & Independent (new) &  0.1 & 0.5 & 62.5 & 81.5   \\ \hline
\multirow{5}{*}{\begin{tabular}[c]{@{}c@{}}Res50-100\% \\ v.s. \\ Res18-50\%\end{tabular}} & $L2$ loss~\cite{yu2019learning} & 13.0 & 32.8 & 43.8 &64.4 \\
& Asymmetric Triplet~\cite{budnik2020asymmetric} & 42.9 & 63.3 & 53.7 & 74.3 \\
& BCT~\cite{shen2020towards} & 42.2 & 65.5 & 55.6 & 76.6 \\ 
&  Dual-Tuning (Ours) & \textbf{46.0} & \textbf{68.8}  & \textbf{62.2} & \textbf{81.5 }\\ \hline
 \end{tabular}}
\vspace{-12pt}
\end{table}

\subsection{Discussion}
In this section, we first perform the ablation study, then explore the feature compatibility under different supervision losses and model architectures. 
Without additional explanation, both the new and old models use ResNet18 backbone, and are supervised by softmax (cross-entropy) loss. 

\textbf{Ablation study.}
Our method contains two parts: a compatible prototype loss and a mutual structural regularization. 
In Table~\ref{tab:aba}, only using the compatible prototype loss can get 80.72\% mAP on the self-test, and even surpasses the independently trained new model 80.43\% mAP (see Table~\ref{tab:market-result}). 
This can be attributed to that prototype loss can exploit the prototype-based manifold information in old embedding space, and meanwhile minimize the impact of noise and disordered distribution derived from the old space. 
Moreover, with additional mutual structural regularization (Mutual-Reg), Dual-Tuning can further boost the cross-test performance, and also obtain good performance on the self-test setting. 

For the ablation study of compatible prototype loss, incorporating new prototypes for a mixed retrieval can achieve better representation than only using the old prototypes (Center-based Prototype Loss). 
Besides, for the mutual structural regularization, the O2N-Reg in Table~\ref{tab:aba} indicates using the old embedding features to train the new head (\textit{i.e.}, $\phi_O$ + $h_N$), and the N2O-Reg uses the old classifier to supervise the new embedding training (\textit{i.e.}, $\phi_N$ + $h_O$). 
The results demonstrate the mutual structural regularization can achieve better performance than only using N2O-Reg or O2N-Reg. 
 
 \begin{table}[t] 
\vspace{-5pt}
\caption{Ablation study on the Market1501 dataset. }
\vspace{-5pt}
\label{tab:aba}
\centering \footnotesize
\renewcommand\arraystretch{1}
\resizebox{1\columnwidth}{!}{
\begin{tabular}{c  c c c  c cc } \\
\hline
  & \multicolumn{3}{c}{Cross-test} & \multicolumn{3}{c}{Self-test}  \\ 
 Method  & mAP   & Top-1 & Top-5 & mAP  & Top-1 & Top-5 \\ \hline 
Center-based Prototype& 67.92 & 85.72 &94.72 & {80.63} & {91.95} & 96.73 \\
Compatible Prototype & 68.12 & 86.10 &94.85 & 80.72 & \textbf{92.30} & 97.21 \\ \hline
 O2N-Reg & 63.84 & 84.29 & 94.12 &  78.86 & 91.32 & 97.26\\
 N2O-Reg & 66.28 & 85.45 & 94.60 & 77.59 & 91.27 & 97.18\\
 Mutual-Reg & 66.82 & 85.60 & 94.69 & 78.81 & 91.38 &97.25  \\
\hline
 Dual-Tuning &  \textbf{69.53} & \textbf{86.67} & \textbf{95.04}  & \textbf{80.75} & 92.04 & \textbf{97.33} \\ \hline 
 \end{tabular}}
 \vspace{-12pt}
\end{table}

\begin{figure}[h]
\centering
\vspace{-10pt}
  \includegraphics[width=0.65\linewidth]{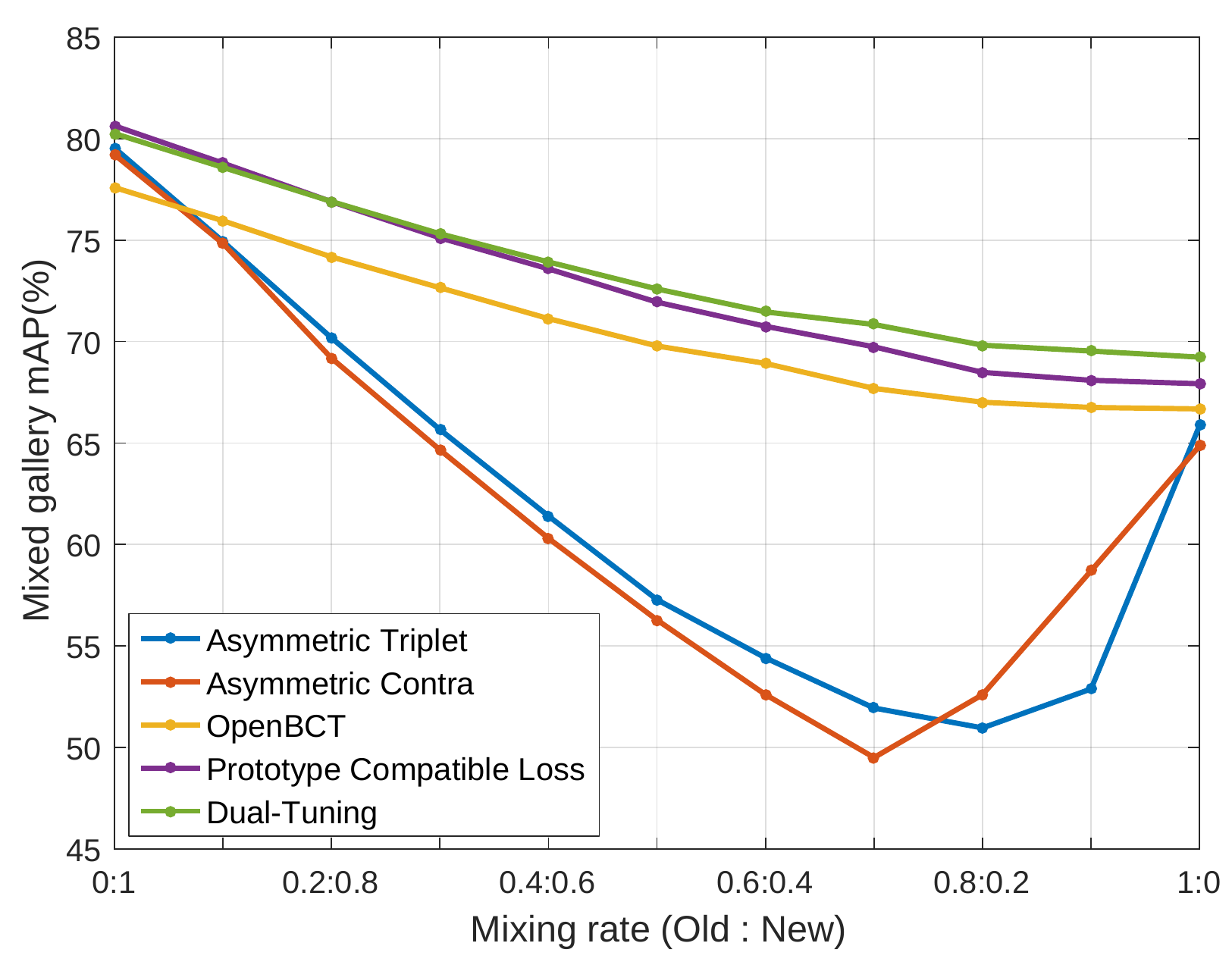}
  \caption{Performances of different methods under mixed dataset.}
    \label{fig:mix}
\vspace{-5pt}
\end{figure}

\textbf{Performances on the mixed gallery.}
In the gallery, there exist the old features and continuously added new features. 
To simulate this scenario, a hybrid gallery retrieval experiment is performed. 
As shown in Fig.~\ref{fig:mix}, we change the ratio of the old and new features in the gallery and report the results of different methods. 
The performances of asymmetric triplet loss and contrastive loss~\cite{budnik2020asymmetric} extremely degrade in the mixed gallery settings. 
Under the setting of 80\% old data and 20\% new data in the gallery, asymmetric triplet can only achieve 50.96\% mAP, which is much inferior to our 70.02\% mAP. Such great performance advantages strongly demonstrate the robustness of our method.


\textbf{Results of different supervision losses.}
Different supervision losses will generate different feature distributions. We test using softmax and triplet loss~\cite{hadsell2006dimensionality} for the old model training, and adopting softmax, softmax+triplet \cite{hadsell2006dimensionality}, and circle loss \cite{sun2020circle} for the new model. 
As shown in Table~\ref{tab:loss}, BCT~\cite{shen2020towards} works poorly when the old model is supervised by triplet loss without a classifier. 
Additionally, BCT suffers huge performance drops in circle v.s. softmax setting, since the circle loss has a different classification mechanism from softmax. 
Differently, our Dual-Tuning achieves impressive results, and beats it by a remarkable margin. 
The superior results indicate the designed compatible prototype loss can mitigate the affection by the distribution discrepancy caused by different supervision losses. 
\begin{table}[t] 
\vspace{-5pt}
\caption{Testing different loss functions on the Market1501. }
\vspace{-5pt}
\label{tab:loss}
\centering \footnotesize
\renewcommand\arraystretch{1}
\resizebox{1\columnwidth}{!}{
\begin{tabular}{c c c  c c c cc } \\
\hline
& 
  \multicolumn{2}{c}{Settings} & \multicolumn{2}{c}{Cross-test} &  \multicolumn{2}{c}{Self-test} \\ 
 Methods  & new model& old model& mAP   & Top-1 & mAP   & Top-1 \\ \hline 
\multicolumn{7}{c}{Independently trained model} \\ \hline
- & - & Soft-50\% & - & - & 61.49 & 82.36  \\
- & - & Trip-50\% & - & - & 57.63 & 78.30 \\
- & Soft-100\% &- &  - & - & 80.43 & 92.31   \\
- & Soft+trip-100\%&- & - & - & 81.08 & 92.70  \\
- & Circle-100\% &- & - & - & 80.76 & 91.28 \\
\hline
\multicolumn{7}{c}{Compatible learning model} \\
\hline
BCT~\cite{shen2020towards} & Soft+trip-100\% & Soft-50\% & 64.45 & 84.23 & 80.91 & 92.20 \\ 
Dual-Tuning & Soft+trip-100\% & Soft-50\% & \textbf{69.47} & \textbf{86.31} & \textbf{81.27} & \textbf{92.62} \\ \hline
BCT~\cite{shen2020towards} & Soft-100\% & Trip-50\%  & - & - & - & - \\
Dual-Tuning & Soft-100\% & Trip-50\%  & \textbf{66.52} & \textbf{84.89} &  \textbf{80.47} & \textbf{92.19} \\ \hline
BCT~\cite{shen2020towards} & Circle-100\% & Soft-50\% & 49.97 & 72.45 & 67.92 & 86.21 \\
Dual-Tuning & Circle-100\% & Soft-50\% & \textbf{64.01} & \textbf{83.90} & \textbf{80.52} & \textbf{91.89} \\ \hline 
 \end{tabular}}
\vspace{-5pt}
\end{table}

\begin{table}[t] 
\caption{Dual-Tuning with different architectures on Market1501. }
\vspace{-5pt}
\label{tab:architecture}
\centering \footnotesize
\renewcommand\arraystretch{0.9}
\resizebox{0.95\columnwidth}{!}{
\begin{tabular}{c c  c ccc c c } \\
\hline
  \multicolumn{2}{c}{Settings} & \multicolumn{2}{c}{Cross-test} &  \multicolumn{2}{c}{Self-test} \\ 
 new model & old model & mAP  & Top-1 & mAP   & Top-1 \\ \hline 
\multicolumn{6}{c}{Independently trained model} \\ \hline
-  & ResNet18-50\% & - & - & 61.49 & 82.36  \\
- & Osnet384-50\% & - & - & 69.95 & 87.00\\
ResNet50-100\%  &-&  - & - & 85.14 & 94.09   \\
Osnet512-100\% &-& - & - & 85.61 & 94.77 \\
ResNeSt50-100\% &-& - & - & 86.84 & 94.81 \\
\hline
\multicolumn{6}{c}{Compatible learning model} \\
\hline
ResNet50-100\% & ResNet18-50\% & 69.98 & 86.74 & 86.41 &  94.42 \\
Osnet512-100\% & ResNet18-50\% & 69.07 & 85.9 & 85.77 &  94.69 \\
ResNeSt50-100\% & ResNet18-50\% & 70.10 & 86.61 & 86.85 & 94.60 \\
Osnet512-100\% & Osnet384-50\% & 73.76 & 89.19 & 85.15 & 94.88 \\
ResNeSt50-100\% & Osnet384-50\% & 73.99 & 88.69 & 86.49 & 94.83 \\
\hline
 \end{tabular}}
\vspace{-12pt}
\end{table}

\textbf{Results of different model architectures.}
We also conduct experiments that the new model with different architectures and feature dimensions from the old model.
First, we test the new model with ResNet50 of 2048 feature dimension, and old model with ResNet18 of 512 feature dimension. 
As shown in Table \ref{tab:market-result}, \ref{tab:Place365-result} and \ref{tab:ImageNet-result}, 
during retrieval with old gallery features, using new ResNet50 feature as query can achieve better performance than the old ResNet18 feature. 

More results of different architectures are shown in Table \ref{tab:architecture}. We test 5 different architectures, including the extremely lightweight networks Osnet384 and Osnet512~\cite{zhou2019omni}, the latest split-attention network ResNeSt50~\cite{zhang2020resnest}, and the widely used networks ResNet18 and ResNet50. The feature dimensions for Osnet384, Osnet512, and ResNeSt50 are 384, 512, and 2048, respectively. 
We use zero padding \cite{shen2020towards} to align features with different dimensions.
Under different architectures, the cross-test between new and old models can also achieve impressive performances. 

\textbf{Towards multi-model and sequential compatibility.} 
We simulate the scene with three model versions on Market1501 dataset. 
These three models are sequentially updated using 25\%, 50\%, and 100\% training data, respectively. 
The version 2 model $\phi_2$ is compatible to version 1 $\phi_1$. The version 3 model $\phi_3$ only conducts supervision from $\phi_2$ to achieve compatibility. The sequential compatible results are shown in Table \ref{tab:seq}. We can find that, $\phi_3$ can achieve feature compatibility with $\phi_1$, even $\phi_1$ is not directly involved in training $\phi_3$. Compared with BCT, we can better solve the sequential updated models, \textit{e.g.}, 60.51\% v.s. 53.64\% mAP on the cross-test between $\phi_3$ and $\phi_1$.  

\begin{table}[t] 
\vspace{-5pt}
\caption{Testing multi-model compatibility on the Market1501. }
\vspace{-5pt}
\label{tab:seq}
\centering \footnotesize
\renewcommand\arraystretch{0.98}
\resizebox{1\columnwidth}{!}{
\begin{tabular}{c c c c  c c c c } \\
\hline
\multicolumn{3}{c}{Settings} & \multicolumn{2}{c}{Cross-test} &  \multicolumn{2}{c}{Self-test} \\ 
Methods & new model & old model  & mAP   & Top-1 & mAP  & Top-1 \\ \hline 
\multicolumn{7}{c}{Independently trained model} \\ \hline 
- & $\phi_1$ & -  &  - & - & 52.44 & 75.36 \\
- & $\phi_2$ & - & - & - & 61.49 & 82.36 \\
- & $\phi_3$ & - & - & - & 80.43 & 92.31 \\ \hline 
\multicolumn{7}{c}{Compatible learning model} \\ \hline 
BCT~\cite{shen2020towards} & $\phi_2$ & $\phi_1$ &51.01 & 73.40 & 61.26 & 82.08  \\
Dual-Tuning & $\phi_2$ & $\phi_1$ &   \textbf{ 53.94 }&  \textbf{ 76.57} &  \textbf{ 63.58} & \textbf{  83.14} \\ \hline
BCT~\cite{shen2020towards}& $\phi_3$ & $\phi_1$ & 53.64 & 76.72  &  79.25 & 91.43 \\
Dual-Tuning & $\phi_3$ & $\phi_1$ & \textbf{ 60.51} & \textbf{81.65} & \textbf{80.54} & \textbf{92.31} \\  \hline
BCT~\cite{shen2020towards} & $\phi_3$ & $\phi_2$ & 66.66 & 86.37 & 79.25 & 91.43 \\
Dual-Tuning & $\phi_3$ & $\phi_2$ & \textbf{69.82} & \textbf{87.50} & \textbf{80.54} & \textbf{92.31} \\
\hline
 \end{tabular}}
\vspace{-5pt}
\end{table}

\textbf{Center-based asymmetric loss.}
After obtaining the old centers, we can also conduct an asymmetric center loss, whose triplet unit is $(\phi_N(x^c), m_o^c, m_o^c{}')$. 
The positive representation is the old prototype $m_o^c$ with the same class as $x^c$, and the negative is an old prototype of other classes $m_o^c{}'$. As shown in Table \ref{tab:aymcenter}, the center-based asymmetric loss can achieve better performance than triplet loss, while worse than our compatible prototype loss. The results further demonstrate the necessity of using the whole structure information in embedding space.

\begin{table}[t] 
\caption{Comparison between metric-based compatible methods. }
\vspace{-5pt}
\label{tab:aymcenter}
\centering \footnotesize
\renewcommand\arraystretch{1}
\resizebox{1\columnwidth}{!}{
\begin{tabular}{c c c c  c c c c } \\
\hline
&& \multicolumn{2}{c}{Cross-test} &  \multicolumn{2}{c}{Self-test} \\ 
Dataset & Method & mAP   & Top-1 & mAP  & Top-1 \\ \hline 
\multirow{3}{*}{MSMT17} & Asymmtric Triplet~\cite{budnik2020asymmetric} &  26.60 & 54.29 & 40.74 & 68.32   \\
& Asymmtric Center &  30.08 & 56.97  & 41.78 & 68.24     \\
& Compatible Prototype & \textbf{32.08} & \textbf{59.70} & \textbf{42.17} & \textbf{70.06} \\ \hline
\multirow{3}{*}{Market1501} & Asymmtric Triplet~\cite{budnik2020asymmetric} &  65.86 & 83.85 & 79.67 & 91.28    \\
& Asymmtric Center & 67.43  & 85.30  & 79.59  &  91.24    \\
& Compatible Prototype & \textbf{68.12} & \textbf{86.10} & \textbf{80.72} & \textbf{92.30} \\
\hline
 \end{tabular}}
\vspace{-12pt}
\end{table}


\section{Conclusion}
In this paper, we propose a Dual-Tuning method to achieve a global optimization for feature compatible learning. Dual-Tuning contains a compatible prototype loss for an explicit embedding feature alignment, and a mutual structural regularization by component interoperation. We conduct experiments on four datasets, including the person ReID datasets and million-scale image classification datasets. Compared with other methods, we can achieve superior compatible performance in mixed gallery scenarios, and can better deal with the distribution discrepancy between new and old embedding spaces. 

{\small
\bibliographystyle{ieee_fullname}
\bibliography{egbib}

\begin{thebibliography}{10}\itemsep=-1pt

\bibitem{budnik2020asymmetric}
Mateusz Budnik and Yannis Avrithis.
\newblock Asymmetric metric learning for knowledge transfer.
\newblock {\em arXiv preprint arXiv:2006.16331}, 2020.

\bibitem{deng2009ImageNet}
Jia Deng, Wei Dong, Richard Socher, Li-Jia Li, Kai Li, and Li Fei-Fei.
\newblock Imagenet: A large-scale hierarchical image database.
\newblock In {\em 2009 IEEE conference on computer vision and pattern
  recognition}, pages 248--255. Ieee, 2009.

\bibitem{deng2019arcface}
Jiankang Deng, Jia Guo, Niannan Xue, and Stefanos Zafeiriou.
\newblock Arcface: Additive angular margin loss for deep face recognition.
\newblock In {\em Proceedings of the IEEE/CVF Conference on Computer Vision and
  Pattern Recognition}, pages 4690--4699, 2019.

\bibitem{fu2019self}
Yang Fu, Yunchao Wei, Guanshuo Wang, Yuqian Zhou, Honghui Shi, and Thomas~S
  Huang.
\newblock Self-similarity grouping: A simple unsupervised cross domain
  adaptation approach for person re-identification.
\newblock In {\em Proceedings of the IEEE International Conference on Computer
  Vision}, pages 6112--6121, 2019.

\bibitem{fukumizu2009kernel}
Kenji Fukumizu, Arthur Gretton, Gert~R Lanckriet, Bernhard Sch{\"o}lkopf, and
  Bharath~K Sriperumbudur.
\newblock Kernel choice and classifiability for rkhs embeddings of probability
  distributions.
\newblock In {\em Advances in neural information processing systems}, pages
  1750--1758, 2009.

\bibitem{ganin2015unsupervised}
Yaroslav Ganin and Victor Lempitsky.
\newblock Unsupervised domain adaptation by backpropagation.
\newblock In {\em International conference on machine learning}, pages
  1180--1189. PMLR, 2015.

\bibitem{gordo2016deep}
Albert Gordo, Jon Almaz{\'a}n, Jerome Revaud, and Diane Larlus.
\newblock Deep image retrieval: Learning global representations for image
  search.
\newblock In {\em European conference on computer vision}, pages 241--257.
  Springer, 2016.

\bibitem{hadsell2006dimensionality}
Raia Hadsell, Sumit Chopra, and Yann LeCun.
\newblock Dimensionality reduction by learning an invariant mapping.
\newblock In {\em 2006 IEEE Computer Society Conference on Computer Vision and
  Pattern Recognition (CVPR'06)}, volume~2, pages 1735--1742. IEEE, 2006.

\bibitem{he2020momentum}
Kaiming He, Haoqi Fan, Yuxin Wu, Saining Xie, and Ross Girshick.
\newblock Momentum contrast for unsupervised visual representation learning.
\newblock In {\em Proceedings of the IEEE/CVF Conference on Computer Vision and
  Pattern Recognition}, pages 9729--9738, 2020.

\bibitem{he2016deep}
Kaiming He, Xiangyu Zhang, Shaoqing Ren, and Jian Sun.
\newblock Deep residual learning for image recognition.
\newblock In {\em Proceedings of the IEEE conference on computer vision and
  pattern recognition(CVPR)}, pages 770--778, 2016.

\bibitem{hinton2015distilling}
Geoffrey Hinton, Oriol Vinyals, and Jeff Dean.
\newblock Distilling the knowledge in a neural network.
\newblock {\em arXiv preprint arXiv:1503.02531}, 2015.

\bibitem{hu2019towards}
Jie Hu, Rongrong Ji, Hong Liu, Shengchuan Zhang, Cheng Deng, and Qi Tian.
\newblock Towards visual feature translation.
\newblock In {\em Proceedings of the IEEE/CVF Conference on Computer Vision and
  Pattern Recognition}, pages 3004--3013, 2019.

\bibitem{Kalayeh2018CVPR}
Mahdi~M. Kalayeh, Emrah Basaran, Muhittin Gökmen, Mustafa~E. Kamasak, and
  Mubarak Shah.
\newblock Human semantic parsing for person re-identification.
\newblock In {\em Proceedings of the IEEE Conference on Computer Vision and
  Pattern Recognition (CVPR)}, June 2018.

\bibitem{lou2019veri}
Yihang Lou, Yan Bai, Jun Liu, Shiqi Wang, and Lingyu Duan.
\newblock Veri-wild: A large dataset and a new method for vehicle
  re-identification in the wild.
\newblock In {\em IEEE Conference on Computer Vision and Pattern Recognition},
  pages 3235--3243, 2019.

\bibitem{luo2019strong}
Hao Luo, Wei Jiang, Youzhi Gu, Fuxu Liu, Xingyu Liao, Shenqi Lai, and Jianyang
  Gu.
\newblock A strong baseline and batch normalization neck for deep person
  re-identification.
\newblock {\em IEEE Transactions on Multimedia}, 2019.

\bibitem{pei2018multi}
Zhongyi Pei, Zhangjie Cao, Mingsheng Long, and Jianmin Wang.
\newblock Multi-adversarial domain adaptation.
\newblock In {\em Proceedings of the AAAI Conference on Artificial
  Intelligence}, volume~32, 2018.

\bibitem{russakovsky2015imagenet}
Olga Russakovsky, Jia Deng, Hao Su, Jonathan Krause, Sanjeev Satheesh, Sean Ma,
  Zhiheng Huang, Andrej Karpathy, Aditya Khosla, Michael Bernstein, et~al.
\newblock Imagenet large scale visual recognition challenge.
\newblock {\em International journal of computer vision}, 115(3):211--252,
  2015.

\bibitem{shen2018wasserstein}
Jian Shen, Yanru Qu, Weinan Zhang, and Yong Yu.
\newblock Wasserstein distance guided representation learning for domain
  adaptation.
\newblock In {\em Proceedings of the AAAI Conference on Artificial
  Intelligence}, volume~32, 2018.

\bibitem{shen2020towards}
Yantao Shen, Yuanjun Xiong, Wei Xia, and Stefano Soatto.
\newblock Towards backward-compatible representation learning.
\newblock In {\em Proceedings of the IEEE/CVF Conference on Computer Vision and
  Pattern Recognition}, pages 6368--6377, 2020.

\bibitem{snell2017prototypical}
Jake Snell, Kevin Swersky, and Richard Zemel.
\newblock Prototypical networks for few-shot learning.
\newblock In {\em Advances in neural information processing systems}, pages
  4077--4087, 2017.

\bibitem{sun2020circle}
Yifan Sun, Changmao Cheng, Yuhan Zhang, Chi Zhang, Liang Zheng, Zhongdao Wang,
  and Yichen Wei.
\newblock Circle loss: A unified perspective of pair similarity optimization.
\newblock {\em arXiv preprint arXiv:2002.10857}, 2020.

\bibitem{Sun2018ECCV}
Yifan Sun, Liang Zheng, Yi Yang, Qi Tian, and Shengjin Wang.
\newblock Beyond part models: Person retrieval with refined part pooling (and a
  strong convolutional baseline).
\newblock In {\em Proceedings of the European Conference on Computer Vision
  (ECCV)}, September 2018.

\bibitem{tang2019pamtri}
Zheng Tang, Milind Naphade, Stan Birchfield, and et al.
\newblock Pamtri: Pose-aware multi-task learning for vehicle re-identification
  using highly randomized synthetic data.
\newblock In {\em IEEE International Conference on Computer Vision}, pages
  211--220, 2019.

\bibitem{Tian2018CVPR}
Maoqing Tian, Shuai Yi, Hongsheng Li, Shihua Li, Xuesen Zhang, Jianping Shi,
  Junjie Yan, and Xiaogang Wang.
\newblock Eliminating background-bias for robust person re-identification.
\newblock In {\em Proceedings of the IEEE Conference on Computer Vision and
  Pattern Recognition (CVPR)}, June 2018.

\bibitem{wang2020unified}
Chien-Yi Wang, Ya-Liang Chang, Shang-Ta Yang, Dong Chen, and Shang-Hong Lai.
\newblock Unified representation learning for cross model compatibility.
\newblock {\em arXiv preprint arXiv:2008.04821}, 2020.

\bibitem{wang2018cosface}
Hao Wang, Yitong Wang, Zheng Zhou, Xing Ji, Dihong Gong, Jingchao Zhou, Zhifeng
  Li, and Wei Liu.
\newblock Cosface: Large margin cosine loss for deep face recognition.
\newblock In {\em Proceedings of the IEEE conference on computer vision and
  pattern recognition}, pages 5265--5274, 2018.

\bibitem{wang2020prototype}
Zijian Wang, Yadan Luo, Zi Huang, and Mahsa Baktashmotlagh.
\newblock Prototype-matching graph network for heterogeneous domain adaptation.
\newblock In {\em Proceedings of the 28th ACM International Conference on
  Multimedia}, pages 2104--2112, 2020.

\bibitem{Wei2017Person}
Longhui Wei, Shiliang Zhang, Wen Gao, and Qi Tian.
\newblock Person transfer gan to bridge domain gap for person
  re-identification.
\newblock In {\em Proceedings of the IEEE Conference on Computer Vision and
  Pattern Recognition (CVPR)}, pages 79--88, 2018.

\bibitem{wen2016discriminative}
Yandong Wen, Kaipeng Zhang, Zhifeng Li, and Yu Qiao.
\newblock A discriminative feature learning approach for deep face recognition.
\newblock In {\em Proceedings of the European Conference on Computer Vision
  (ECCV)}, pages 499--515, 2016.

\bibitem{wu2018unsupervised}
Zhirong Wu, Yuanjun Xiong, Stella~X Yu, and Dahua Lin.
\newblock Unsupervised feature learning via non-parametric instance
  discrimination.
\newblock In {\em Proceedings of the IEEE Conference on Computer Vision and
  Pattern Recognition}, pages 3733--3742, 2018.

\bibitem{xu2020cross}
Minghao Xu, Hang Wang, Bingbing Ni, Qi Tian, and Wenjun Zhang.
\newblock Cross-domain detection via graph-induced prototype alignment.
\newblock In {\em Proceedings of the IEEE/CVF Conference on Computer Vision and
  Pattern Recognition}, pages 12355--12364, 2020.

\bibitem{yan2017mind}
Hongliang Yan, Yukang Ding, Peihua Li, Qilong Wang, Yong Xu, and Wangmeng Zuo.
\newblock Mind the class weight bias: Weighted maximum mean discrepancy for
  unsupervised domain adaptation.
\newblock In {\em Proceedings of the IEEE Conference on Computer Vision and
  Pattern Recognition}, pages 2272--2281, 2017.

\bibitem{yao2020knowledge}
Anbang Yao and Dawei Sun.
\newblock Knowledge transfer via dense cross-layer mutual-distillation.
\newblock In {\em European Conference on Computer Vision}, pages 294--311.
  Springer, 2020.

\bibitem{yim2017gift}
Junho Yim, Donggyu Joo, Jihoon Bae, and Junmo Kim.
\newblock A gift from knowledge distillation: Fast optimization, network
  minimization and transfer learning.
\newblock In {\em Proceedings of the IEEE Conference on Computer Vision and
  Pattern Recognition}, pages 4133--4141, 2017.

\bibitem{yu2019learning}
Lu Yu, Vacit~Oguz Yazici, Xialei Liu, Joost van~de Weijer, Yongmei Cheng, and
  Arnau Ramisa.
\newblock Learning metrics from teachers: Compact networks for image embedding.
\newblock In {\em Proceedings of the IEEE/CVF Conference on Computer Vision and
  Pattern Recognition}, pages 2907--2916, 2019.

\bibitem{zagoruyko2016paying}
Sergey Zagoruyko and Nikos Komodakis.
\newblock Paying more attention to attention: Improving the performance of
  convolutional neural networks via attention transfer.
\newblock {\em arXiv preprint arXiv:1612.03928}, 2016.

\bibitem{zhang2020resnest}
Hang Zhang, Chongruo Wu, Zhongyue Zhang, Yi Zhu, Zhi Zhang, Haibin Lin, Yue
  Sun, Tong He, Jonas Mueller, R Manmatha, et~al.
\newblock Resnest: Split-attention networks.
\newblock {\em arXiv preprint arXiv:2004.08955}, 2020.

\bibitem{zheng2015scalable}
Liang Zheng, Liyue Shen, Lu Tian, Shengjin Wang, Jingdong Wang, and Qi Tian.
\newblock Scalable person re-identification: A benchmark.
\newblock In {\em IEEE International Conference on Computer Vision}, pages
  1116--1124, 2015.

\bibitem{zhong2020bi}
Jincheng Zhong, Ximei Wang, Zhi Kou, Jianmin Wang, and Mingsheng Long.
\newblock Bi-tuning of pre-trained representations.
\newblock {\em arXiv preprint arXiv:2011.06182}, 2020.

\bibitem{zhou2017places}
Bolei Zhou, Agata Lapedriza, Aditya Khosla, Aude Oliva, and Antonio Torralba.
\newblock Places: A 10 million image database for scene recognition.
\newblock {\em IEEE transactions on pattern analysis and machine intelligence},
  40(6):1452--1464, 2017.

\bibitem{zhou2019omni}
Kaiyang Zhou, Yongxin Yang, Andrea Cavallaro, and Tao Xiang.
\newblock Omni-scale feature learning for person re-identification.
\newblock In {\em Proceedings of the IEEE/CVF International Conference on
  Computer Vision}, pages 3702--3712, 2019.

\bibitem{zou2018unsupervised}
Yang Zou, Zhiding Yu, BVK Kumar, and Jinsong Wang.
\newblock Unsupervised domain adaptation for semantic segmentation via
  class-balanced self-training.
\newblock In {\em Proceedings of the European conference on computer vision
  (ECCV)}, pages 289--305, 2018.

\end{thebibliography}
}

\end{document}